\documentclass[unnumsec,webpdf,contemporary,large]{oup-authoring-template} 

\usepackage{graphicx}
\usepackage{booktabs}
\usepackage{bbm}
\usepackage{hyperref}

\usepackage{setspace}

\usepackage{makecell}

\graphicspath{{Fig/}}

\theoremstyle{thmstyleone}%
\theoremstyle{thmstyletwo}%
\theoremstyle{thmstylethree}%

\begin{document}

\journaltitle{Journal Title Here}
\DOI{DOI HERE}
\copyrightyear{2022}
\pubyear{2019}
\access{Advance Access Publication Date: Day Month Year}
\appnotes{Paper}

\firstpage{1}


\title{To Reason or Not to: Selective Chain-of-Thought in Medical Question Answering}

\author[1,2]{Zaifu Zhan, MEng}
\author[2]{Min Zeng, PhD}
\author[2]{Shuang Zhou, PhD}
\author[2]{Yiran Song, PhD}
\author[2]{Xiaoyi Chen, PhD}
\author[2]{Yu Hou, PhD}
\author[3]{Yifan Wu, MS}
\author[4]{Yang Ruan, MS}
\author[2,*]{Rui Zhang, PhD}

\authormark{Zaifu Zhan et al.}

\address[1]{\orgdiv{Department of Electrical and Computer Engineering}, 
            \orgname{University of Minnesota}, 
            \orgaddress{\street{200 Union St SE}, \postcode{55455}, \state{Minneapolis, MN}, \country{United States}}}

\address[2]{\orgdiv{Division of Computational Health Sciences, Department of Surgery}, 
            \orgname{University of Minnesota}, 
            \orgaddress{\street{420 Delaware St SE}, \postcode{55455}, \state{Minneapolis, MN}, \country{United States}}}

\address[3]{\orgdiv{Department of Computer Science and Engineering}, 
            \orgname{University of Minnesota}, 
            \orgaddress{\street{200 Union St SE}, \postcode{55455}, \state{Minneapolis, MN}, \country{United States}}}
\address[4]{\orgdiv{Division of Biostatistics and Health Data Science, School of Public Health}, 
            \orgname{University of Minnesota}, 
            \orgaddress{\street{2221 University Ave SE}, \postcode{55414}, \state{Minneapolis, MN}, \country{United States}}}

\corresp[$\ast$]{Corresponding author:
Dr. Rui Zhang, PhD, Division of Computational Health Sciences, Department of Surgery, University of Minnesota, Office: D528 Mayo building, 420 Delaware St SE, Minneapolis, MN 55455,
\href{mailto:ruizhang@umn.edu}{ruizhang@umn.edu}, Office Phone: 612-626-4209\\
\\
Full paper word count: 2564\\
Abstract word count: 187
}

\abstract{
\doublespacing
\textbf{Objective:} 
To improve the efficiency of medical question answering (MedQA) with large language models (LLMs) by avoiding unnecessary reasoning while maintaining accuracy.\\
\textbf{Methods:} 
We propose Selective Chain-of-Thought (Selective CoT), an inference-time strategy that first predicts whether a question requires reasoning and generates a rationale only when needed. 
Two open-source LLMs (Llama-3.1-8B and Qwen-2.5-7B) were evaluated on four biomedical QA benchmarks—HeadQA, MedQA-USMLE, MedMCQA, and PubMedQA. 
Metrics included accuracy, total generated tokens, and inference time.\\
\textbf{Results:} 
Selective CoT reduced inference time by 13–45\% and token usage by 8–47\% with minimal accuracy loss ($\leq$4\%). 
In some model–task pairs, it achieved both higher accuracy and greater efficiency than standard CoT. 
Compared with fixed-length CoT, Selective CoT reached similar or superior accuracy at substantially lower computational cost.\\
\textbf{Discussion:} 
Selective CoT dynamically balances reasoning depth and efficiency by invoking explicit reasoning only when beneficial, reducing redundancy on recall-type questions while preserving interpretability.\\
\textbf{Conclusion:} 
Selective CoT provides a simple, model-agnostic, and cost-effective approach for medical QA, aligning reasoning effort with question complexity to enhance real-world deployability of LLM-based clinical systems.
}

\keywords{Question answering, Large language models, Chain-of-thought, Reasoning, Efficiency}

\maketitle
\doublespacing

\section{Introduction}

Medical question answering (QA)~\cite{jin2022biomedical}, broadly including both clinical and biomedical QA, is an important challenge across medicine and the biomedical sciences~\cite{Hanjie2025benchmarking}, as it demands two competencies: precise factual recall of vast, long-tail biomedical knowledge~\cite{yang2024kg} and structured reasoning~\cite{zhou2025automating} across clinical context, evidence, and distractors. Traditional NLP pipelines (e.g., keyword-matching search, manually crafted feature-based models, machine learning models, or rule-based systems)~\cite{jin2021disease,xiong2024benchmarking,abdallah2020automated} have struggled to meet this dual requirement: they are brittle to paraphrase and domain shift, offer limited compositionality, and often fail to integrate dispersed cues into coherent, stepwise judgments. The advent of large language models (LLMs)~\cite{zhou2025large,zhan2025evaluation,hou2025benchmarking} has materially changed this landscape along the two axes most stressed by MedQA: (1) knowledge coverage and access (via broad pretraining, domain adaptation, and retrieval-augmented generation that improve the memorization of rare entities and guidelines)~\cite{luo2024biomedgpt,lan2025large,pal2024domain,shool2025systematic} and (2) reasoning capacity (via emergent abilities for multistep inference, uncertainty articulation, and resistance to distractors, further amplified by techniques such as chain-of-thought prompting, self-consistency, and tool use)~\cite{wang2025medical,lucas2024reasoning,Li2024MediQ,jeong2024improving}.

Chain-of-thought (CoT) prompting~\cite{Wei2022cot}, in which a model generates intermediate reasoning steps before giving a final answer, has emerged as a strong default for such settings. CoT can improve logicality and interpretability when questions require multi-step reasoning~\cite{lievin2024can,jeon2025comparative}. 
Numerous studies have leveraged CoT prompting to improve LLMs on the MedQA benchmark. For example, Jeon et al.~\cite{jeon2025comparative} conduct a comprehensive evaluation of CoT across both clinical and non-clinical settings, benchmarking GPT- and Gemini-based models. Le et al.~\cite{le2025instruction} tackle contextual medical QA by combining instruction tuning with CoT prompting. Going further, Wang et al.~\cite{wang2025medcot} integrate CoT with retrieval-augmented generation (RAG) to further enhance model performance.
However, always generating long rationales carries nontrivial costs: more output tokens, higher latency, and increased compute expenditure, especially on knowledge-recall questions where explicit reasoning is unnecessary. This accuracy–efficiency trade-off is acute in realistic deployments (education platforms, decision support), where service-level objectives are tied to throughput and responsiveness as much as to accuracy.

To avoid this waste, we introduce Selective Chain-of-Thought (Selective CoT), a simple and effective inference-time routing strategy that first decides whether a question requires explicit reasoning and, only when warranted, generates a rationale before answering. The premise is straightforward: a model capable of reasoning should also be able to recognize when reasoning is needed. In clinical and operational environments, where many queries are recall-oriented, this approach aims to preserve accuracy while reducing token consumption and end-to-end latency by avoiding gratuitous rationales. Conceptually, Selective CoT synthesizes the strengths of standard direct answering (speed) and full CoT (reasoning on hard items), shifting part of orchestration from external control to the model’s own self-selection mechanism.
\begin{figure*}[htbp]
    \centering
    \includegraphics[width=0.9\linewidth]{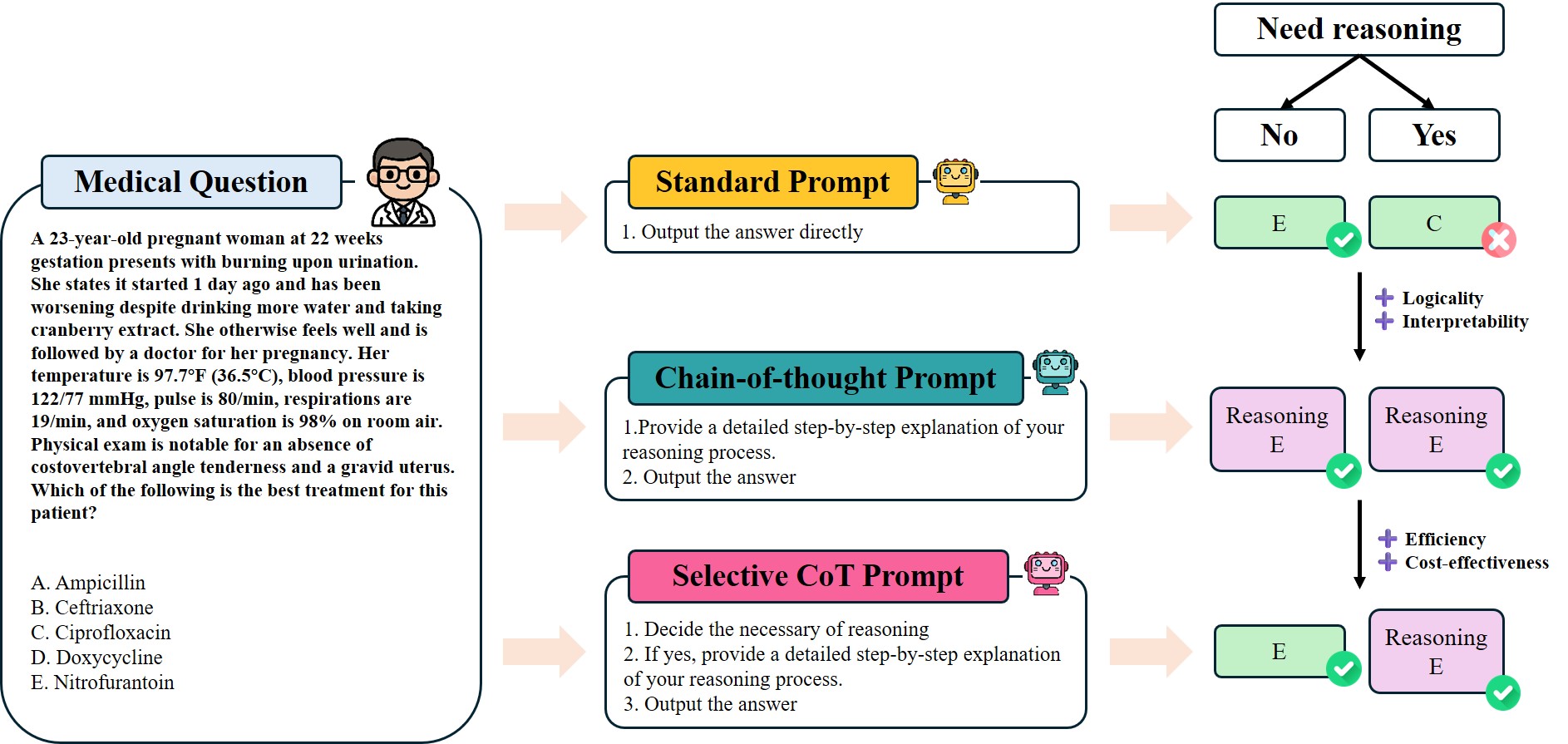}
    \caption{Overview of Standard Prompt, Chain-of-thought, and Selective CoT prompting (ours) on a representative MedQA case. Left: an example clinical vignette (multiple-choice from MedQA-USMLE dataset). Middle: three inference paradigms—Standard Prompt (answer directly), CoT Prompt (produce step-by-step reasoning before the answer), and Selective CoT Prompt (first decide whether reasoning is necessary; if yes, generate CoT, otherwise answer directly). Right: a comparison illustrating that full CoT improves logicality and interpretability when reasoning is required, but wastes compute on recall-type questions; Selective CoT preserves accuracy while improving efficiency and cost-effectiveness by invoking reasoning only when needed.}
    \label{fig:method}
\end{figure*}

To test this idea, we evaluate Selective CoT across four complementary biomedical QA benchmarks across HeadQA, MedQA-USMLE, MedMCQA, and PubMedQA. Together, this suite blends exam-style and evidence-based QA and supports a comprehensive assessment of factual knowledge, cross-lingual robustness, and interpretive judgment. In addition, our experiments use two strong, accessible open-source LLMs, Llama-3.1-8B and Qwen-2.5-7B, and report a three-way evaluation: answer accuracy, total output tokens, and wall-clock inference time. We fix random seeds, cap generation length to accommodate rationales, and disclose environment details to support reproducibility.
Our contributions are summarized as:
\begin{itemize}
    \item To the best of our knowledge, this is among the first studies to explicitly investigate selective reasoning control for improving the efficiency of large language models in medical question answering.
    \item We introduced Selective CoT, a simple, model-agnostic strategy that decides per question whether to generate an explicit rationale, preserving accuracy while reducing tokens and latency relative to always-on CoT.
    \item We provided a comprehensive evaluation. Across HeadQA, MedMCQA, MedQA-USMLE, and PubMedQA with the open-source models Llama-3.1-8B and Qwen-2.5-7B, Selective CoT typically reduces time by 13\% to 45\% and tokens by 8\% to 47\% with only small accuracy changes; in some model–task pairs, it is both faster and more accurate.
\end{itemize}
\section{Methods}

\subsection{Overview of methods}

\subsection{Task and datasets}
\noindent\textbf{Medical QA Task.}
Medical QA is a specialized form of question answering that focuses on understanding, retrieving, and reasoning over medical knowledge to provide accurate and reliable answers to health-related questions. The task requires models to integrate biomedical facts, clinical guidelines, and contextual understanding to answer questions ranging from basic biomedical science to complex clinical diagnosis and treatment scenarios. Unlike general-domain QA, medical QA emphasizes factual correctness, interpretability, and safety, as errors may have significant real-world implications. It serves as a critical benchmark for evaluating the reasoning and domain adaptation capabilities of language models in the healthcare domain, with potential applications in medical education, clinical decision support, and patient-facing information systems.

\noindent\textbf{HeadQA dataset~\cite{headqa}.}
HeadQA is a Spanish multiple-choice medical QA dataset built from official healthcare entrance and specialization exams in Spain. Items span several disciplines (e.g., medicine, nursing, psychology, chemistry, pharmacology, and biology) and reflect authentic exam difficulty. It is commonly used to assess cross-disciplinary medical reasoning and cross-lingual generalization in clinical QA models.

\noindent\textbf{MedQA–USMLE dataset~\cite{jin2021disease}.}
MedQA-USMLE contains multiple-choice questions sourced from the United States Medical Licensing Examination and related professional board materials. The questions emphasize clinical reasoning, pathophysiology, and diagnosis across core rotations (internal medicine, surgery, pediatrics, OB/GYN, psychiatry, etc.). It serves as a benchmark for evaluating a model’s performance on real, high-stakes U.S. medical exam items.

\noindent\textbf{MedMCQA dataset~\cite{pal2022medmcqa}.}
MedMCQA is a large-scale, multi-disciplinary multiple-choice dataset derived from Indian medical entrance and postgraduate exams (e.g., AIIMS, NEET-PG). It covers over 20 subjects and thousands of topics, with carefully designed distractors that require both foundational knowledge and integrated clinical reasoning. The dataset is frequently used to test robustness in multi-subject medical understanding.

\noindent\textbf{PubMedQA dataset~\cite{jin2019pubmedqa}.}
PubMedQA targets biomedical literature question answering, where each instance poses a research-style question answered as \emph{yes/no/maybe}. Models must infer the answer from evidence in the linked abstract, encouraging careful reasoning over study summaries. It is suited for evaluating evidence aggregation and causal/interpretive judgment in biomedical QA.

Taken together, our four datasets constitute a well-rounded benchmark suite. HeadQA contributes exam questions from Spanish and multi-disciplinary coverage; MedQA–USMLE provides high-stakes U.S. licensure items emphasizing clinical reasoning; MedMCQA adds large-scale, multi-subject questions from the Indian exam ecosystem; and PubMedQA complements exam-style QA with literature-grounded inference over biomedical abstracts. This mix spans styles, regions, and knowledge sources (exams vs.\ scientific literature), as well as task formats (multiple-choice vs.\ textual evidence). Consequently, it enables a comprehensive assessment of factual knowledge, cross-source robustness, and evidence-based clinical reasoning—offering both breadth and depth for reliable model evaluation. The dataset statistics and sources are summarized in Table \ref{tab:medqa_datasets}.
\begin{table}[htbp]
\centering
\begin{tabular}{lccccc}
\hline
\textbf{Dataset} & \textbf{Testset} & \textbf{Source}\\
\hline
HeadQA        & 244   & Spanish medical exams \\
MedQA-USMLE & 1,273 & US Medical exam \\
MedMCQA & 4,183 & Indian Medical exam \\
PubMedQA      & 500 & Biomedical literature  \\
\hline
\end{tabular}
\caption{Datasets and their source. We report only test sets, as no training or fine-tuning was performed.}
\label{tab:medqa_datasets}
\end{table}

\subsection{Selective Chain-of-thought}
From the perspective of whether explicit reasoning is required, medical questions can be broadly grouped into (i) knowledge-recall items that can be answered directly from established medical facts (e.g., whether a drug is indicated for a disease), and (ii) reasoning items that involve misleading cues, multi-step calculation, temporal ordering, or constraint composition, as shown in Figure \ref{fig:method}. These categories align with two inference paradigms: standard prompting, which generates an answer directly, and chain-of-thought (CoT) prompting, which first produces intermediate reasoning and then the final answer. While full CoT often matches or exceeds the accuracy of standard prompting, it incurs substantial computational cost and latency, particularly on knowledge-recall questions where explicit reasoning is unnecessary.

To balance accuracy and efficiency, we introduce Selective CoT, as shown in Figure \ref{fig:method}, in which the model adaptively decides at inference time whether to invoke CoT. If the question is judged to require reasoning, CoT is enabled; otherwise, the system answers directly. The core premise is that a model capable of reasoning should also be able to recognize when reasoning is needed. In clinical and real-world settings—where many queries are knowledge-recall—Selective CoT preserves accuracy while reducing token consumption and end-to-end latency by avoiding gratuitous reasoning. The method is conceptually simple and synthesizes the strengths of standard and full-CoT prompting. Looking ahead, as large models’ reasoning capabilities continue to improve, the ability to decide whether to reason will become increasingly important, shifting part of the burden from external orchestration to the model’s own self-selection mechanism.

\subsection{Prompt}
In our experiments, we designed and utilized various prompting strategies to evaluate the reasoning capabilities of the models. 
As shown in Figure \ref{fig:method}, the CoT prompt serves as the baseline, explicitly instructing the models to produce comprehensive and elaborate reasoning processes before providing a final answer. The selective CoT prompt introduces a mechanism for selectively generating reasoning, balancing between detail and efficiency. Additionally, we did experiments with prompts with explicitly specified reasoning lengths: 100 words, 200 words, and so on, where the weights control the level of detail in the reasoning process. These variations allow us to systematically analyze the trade-offs between reasoning depth, computational cost, and answer quality across different prompting settings. 
All prompts used in our experiments, including CoT and selective CoT,
and fixed-length CoT variants, will be released with the code for
reproducibility.

\subsection{Models}
In our experiments, we utilized two open-source large language models: Llama-3.1-8B~\cite{dubey2024llama} and Qwen2.5-7B~\cite{qwen2,qwen2.5}. Both models were selected for their accessibility and strong performance in reasoning tasks, making them suitable for evaluating our framework. The use of open-source models ensures that our experiments can be easily reproduced by others, fostering transparency and enabling further research in selective reasoning and medical question answering.

\subsection{Metrics}
We evaluated the performance of the models using several standard metrics to ensure a comprehensive assessment. Accuracy was used to measure the proportion of correct answers among all predictions. The total number of tokens (computed via tokenizer) and the total inference time were also recorded to analyze the computational efficiency of the models. These metrics provide a holistic view of both the correctness and efficiency of the reasoning and answering processes.
Although we report both token counts and wall-clock time, noting that token usage is a more stable, model-proximal efficiency metric, while latency reflects
end-to-end deployment cost.

\subsection{Experiments}
All experiments were conducted on NVIDIA A100 GPUs with 40 GB of memory to ensure sufficient computational resources for running large language models. We used a temperature of 0.7 for sampling to encourage diverse reasoning paths while maintaining coherence. The maximum generation length was set to 1,024 tokens to accommodate detailed reasoning outputs. For evaluation, we fixed random seeds to ensure reproducibility across runs. The experiments were implemented using Python, with key libraries including PyTorch~\cite{paszke2019pytorch} and Hugging Face Transformers~\cite{wolf2020transformers}, to facilitate efficient model inference and result analysis.

\begin{table*}[htbp]
\centering
\caption{Selective CoT vs. full CoT: accuracy–efficiency trade-offs. Accuracy, total generated tokens, and end-to-end inference time across datasets and models; “Drop(\%)” denotes relative change vs. full CoT (positive means reduction).}
\label{tab:acc_two_cols}
\begin{tabular}{llccccccc}
\toprule
\multicolumn{3}{c}{} & \multicolumn{2}{c}{Acc} & \multicolumn{2}{c}{\# Tokens} & \multicolumn{2}{c}{Time (s)} \\
\cmidrule(lr){4-5}\cmidrule(lr){6-7}\cmidrule(lr){8-9}
Dataset & Model & Method & Score & Drop (\%) & Value & Drop (\%) & Value & Drop (\%) \\
\midrule
\multirow{4}{*}{HeadQA} & \multirow{2}{*}{Llama-3.1-8B} & CoT & 0.7090 &  & 135504 &  & 2695.1428 &  \\
 &  & Selective CoT & 0.6803 & 4.0462 & 136880 & -1.0155 & 2178.3438 & 19.1752 \\
 & \multirow{2}{*}{Qwen2.5-7B} & CoT & 0.6598 &  & 93352 &  & 2574.1979 &  \\
 &  & Selective CoT & 0.7172 & -8.6957 & 75600 & 19.0162 & 2120.5520 & 17.6228 \\
\hline
\multirow{4}{*}{MedMCQA} & \multirow{2}{*}{Llama-3.1-8B} & CoT & 0.5826 &  & 2184424 &  & 45140.2206 &  \\
 &  & Selective CoT & 0.5668 & 2.7082 & 2113419 & 3.2505 & 33378.0168 & 26.0570 \\
 & \multirow{2}{*}{Qwen2.5-7B} & CoT & 0.5252 &  & 1501200 &  & 42031.0936 &  \\
 &  & Selective CoT & 0.5274 & -0.4096 & 1148503 & 23.4943 & 32292.8344 & 23.1692 \\
\hline
\multirow{4}{*}{MedQA-USMLE} & \multirow{2}{*}{Llama-3.1-8B} & CoT & 0.6465 &  & 1034709 &  & 19077.2906 &  \\
 &  & Selective CoT & 0.6402 & 0.9721 & 953726 & 7.8266 & 13487.3329 & 29.3016 \\
 & \multirow{2}{*}{Qwen2.5-7B} & CoT & 0.5711 &  & 636615 &  & 18494.5302 &  \\
 &  & Selective CoT & 0.5460 & 4.4017 & 547417 & 14.0113 & 15517.1266 & 16.0988 \\
\hline
\multirow{4}{*}{PubMedQA} & \multirow{2}{*}{Llama-3.1-8B} & CoT & 0.7480 &  & 357037 &  & 4610.8540 &  \\
 &  & Selective CoT & 0.7280 & 2.6738 & 380567 & -6.5904 & 4004.7968 & 13.1441 \\
 & \multirow{2}{*}{Qwen2.5-7B} & CoT & 0.7320 &  & 207309 &  & 5921.6214 &  \\
 &  & Selective CoT & 0.7280 & 0.5464 & 110907 & 46.5016 & 3284.0856 & 44.5408 \\
\bottomrule
\end{tabular}
\end{table*}
\section{Results}

\subsection{Main results}
As shown in Table~\ref{tab:acc_two_cols}, across four biomedical benchmarks (HeadQA, MedMCQA, MedQA-USMLE, PubMedQA), we compare standard CoT with Selective CoT for Llama-3.1-8B and Qwen2.5-7B. Overall, Selective CoT consistently reduces inference cost—wall-clock time drops by $\sim$13--45\% and output tokens often shrink by $\sim$8--47\%—with only minor accuracy changes (typically within 0--4\%). Notably, Qwen2.5-7B on HeadQA is “faster and more accurate,” improving accuracy by 8.70\% while cutting tokens ($-19.0\%$) and time ($-17.6\%$). On MedMCQA, accuracy shifts are negligible (Llama $-2.71\%$, Qwen $+0.41\%$), yet time and token counts fall by $\sim$23--26\%. On MedQA-USMLE, Selective CoT trades small accuracy declines (Llama $-0.97\%$, Qwen $-4.40\%$) for sizable efficiency gains (time $-16$--$29\%$; tokens $-8$--$14\%$). On PubMedQA, Qwen attains near-parity in accuracy ($-0.55\%$) with the largest efficiency savings (tokens $-46.5\%$, time $-44.5\%$), whereas Llama sees a modest time reduction ($\sim$13\%) but a slight token increase. In sum, Selective CoT typically yields substantial speedups and token savings at minimal accuracy cost, and can even deliver simultaneous gains in both accuracy and efficiency for certain model–task pairs.

\begin{figure*}[htbp]
    \centering
    \includegraphics[width=0.99\linewidth]{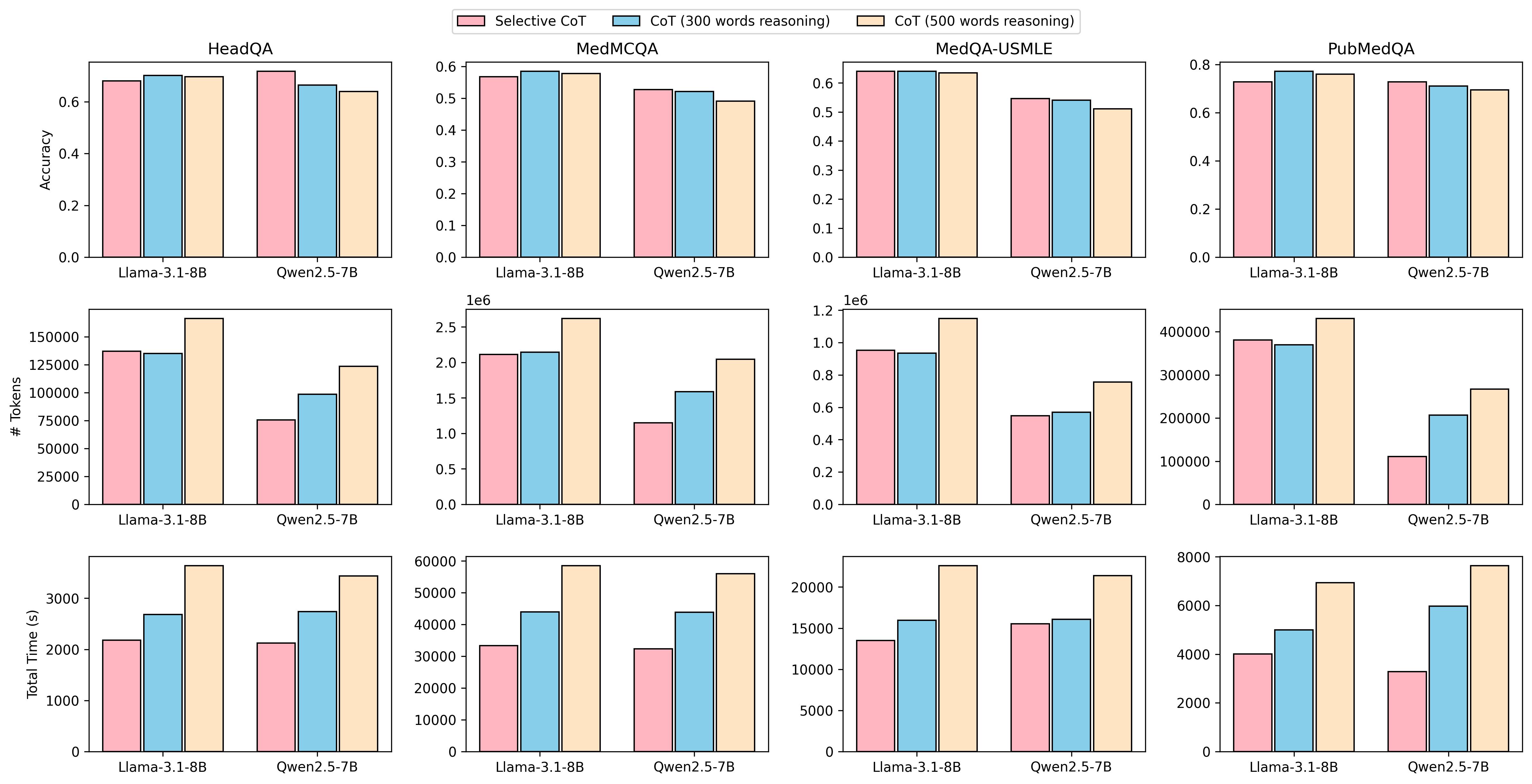}
    \caption{Performance and efficiency comparison of Selective CoT versus fixed-length CoT across four biomedical QA datasets (HeadQA, MedMCQA, MedQA-USMLE, PubMedQA) and two open-source LLMs (Llama-3.1-8B, Qwen2.5-7B). The three rows report, respectively, \emph{Accuracy}, \emph{\#Tokens}, and \emph{Inference Time}. Selective CoT matches or slightly outperforms strong fixed-length CoT baselines while substantially reducing token usage and latency, yielding a superior compute--performance trade-off.}
    \label{fig:diff_words_bar}
\end{figure*}

\subsection{Ablation Study: Reasoning Length vs. Selective CoT}
\begin{figure*}[t]
    \centering
    \includegraphics[width=0.99\linewidth]{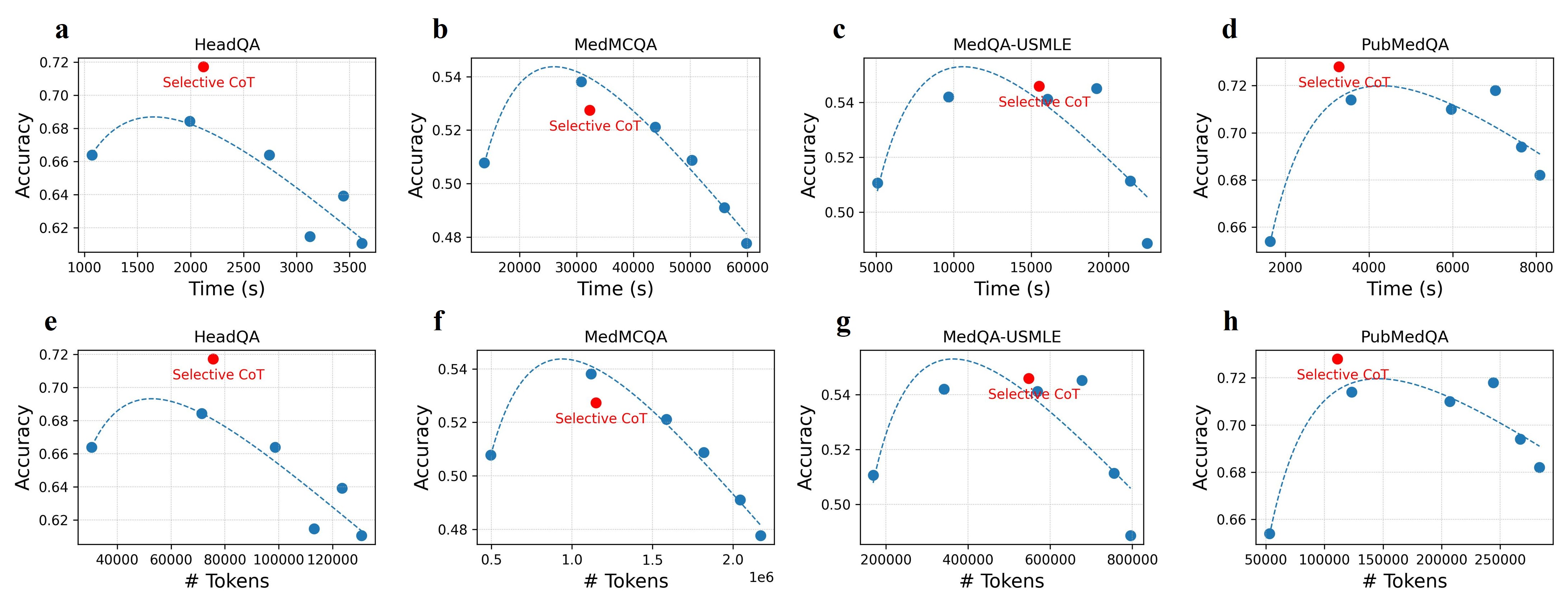}
    \caption{Ablation on reasoning length versus Selective CoT. For fixed-length CoT, we sweep reasoning lengths (e.g., 100--600 words) and fit a dashed quadratic curve to the length--accuracy relationship; Selective CoT is marked as a red point for reference. Accuracy under Selective CoT typically lies near the empirical optimum and often on or above the fitted curve for several datasets, achieving comparable or higher accuracy with fewer tokens and shorter time than long, uniform CoT.}
    \label{fig:diff_words_curves}
\end{figure*}

We compare Selective CoT with explicitly controlled fixed-length CoT. Specifically, for Llama-3.1-8B and Qwen-2.5-7B, we evaluate Selective CoT, CoT with approximately 300 words, and CoT with approximately 500 words (Fig.~\ref{fig:diff_words_bar}). In addition, on Qwen-2.5-7B we sweep the CoT length more finely from 100 to 600 words in 100-word increments and fit a quadratic curve to the “length–performance” relationship, marking Selective CoT as a red point for reference (Fig.~\ref{fig:diff_words_curves}).

Across the four datasets, accuracy under Selective CoT is comparable to that of the 300-word and 500-word CoT baselines; in several cases (HeadQA, USMLE, and PubMedQA), Selective CoT is slightly better, while on MedMCQA it is close to the best fixed-length setting. Crucially, Selective CoT achieves these results with a substantially smaller number of Tokens and shorter inference time (see the second and third rows of Fig.~\ref{fig:diff_words_bar}), yielding a markedly better compute–performance trade-off.

The length sweep reveals a non-monotonic pattern: as the fixed CoT length increases from 100 to roughly 200–300 words, accuracy first rises and then declines (the dashed quadratic peak in Fig.~\ref{fig:diff_words_curves}). Selective CoT lies near this empirical optimum and, on HeadQA, USMLE, and PubMedQA, it sits on or above the fitted CoT curve—achieving equal or higher accuracy with lower cost; on MedMCQA it remains close to the curve’s peak. Overall, Selective CoT forms a superior frontier, maintaining similar (or higher) accuracy while significantly reducing \#Tokens and time by triggering extended reasoning only when needed, rather than enforcing a uniform long reasoning length for all examples.

\section{Discussion}
Question answering in biomedicine underpins clinical education, decision support, and patient safety, where models must deliver both correctness and responsiveness under real-world compute and latency constraints. Medical QA items broadly split into recall-type versus reasoning-dependent questions: full CoT can benefit the latter but often wastes tokens and time on the former. Building on this observation, we propose Selective CoT, which decides per example whether explicit reasoning is necessary, aiming to preserve accuracy while reducing inference cost and improving practical deployability.

Across four biomedical QA benchmarks (HeadQA, MedMCQA, MedQA-USMLE, PubMedQA) and two open-source LLMs (e.g., Llama-3.1-8B, Qwen2.5-7B), Selective CoT consistently reduced inference cost (wall-clock time and output tokens) with only minor accuracy changes relative to full CoT. On HeadQA and PubMedQA, accuracy matched or slightly exceeded fixed-length CoT (e.g., 300/500 tokens), while on MedMCQA and USMLE, accuracy was comparable to the best fixed-length baselines at substantially lower generation length and latency. Overall, Selective CoT yields a better compute--performance trade-off by generating reasoning traces only when they are likely to help.

Further, we compared Selective CoT with fixed-length CoT across a sweep of reasoning lengths and fitted a second-order (quadratic) curve to the length–accuracy relationship. The fitted curve captures the typical rise toward an intermediate optimum followed by saturation or decline. Selective CoT typically lies on or above this empirical curve for most datasets, indicating that example-level on/off routing is more decisive than merely increasing the length of a uniform rationale. In practice, Selective CoT reaches near-peak accuracy without per-task length tuning while using fewer tokens and incurring shorter latency, yielding a more reliable and compute-efficient operating point.

One limitation of this study is that the efficiency metric, wall-clock time, is sensitive to backend factors (hardware, system load, batching, runtime). Token usage is a more stable, model-proximal proxy, but end users ultimately care about latency. Accordingly, we report both tokens and time and disclose environment details; while time can be noisy, it still provides a coarse basis for comparison.

\section{Conclusion}
We introduced Selective CoT, which is an inference-time routing strategy that first decides whether a question requires explicit reasoning and, only when warranted, generates a rationale before answering. Across four biomedical QA benchmarks (HeadQA, MedMCQA, MedQA–USMLE, PubMedQA) and two open LLMs, Selective CoT preserved—or in some cases improved—accuracy while substantially reducing token usage and end-to-end latency compared to always-on CoT. By aligning the amount of reasoning with item difficulty, Selective CoT delivers interpretable rationales where they matter most and lightweight direct answers where they suffice, offering a practical, cost-efficient path for deployable clinical QA systems under real-world budget and responsiveness constraints.

\section{Data Availability}
The four QA datasets we used are all public:
\begin{enumerate}
    \item HeadQA: \href{https://aghie.github.io/head-qa/}{https://aghie.github.io/head-qa/}
    \item MedMCQA:     \href{https://medmcqa.github.io/}{https://medmcqa.github.io/}
    \item MedQA(USMLE): \href{https://github.com/jind11/MedQA}{https://github.com/jind11/MedQA}
    \item PubmedQA: \href{https://pubmedqa.github.io/}{https://pubmedqa.github.io/}
\end{enumerate}

\section{Code Availability}
We will release the code upon the acceptance of the paper.

\section{Acknowledgements}
We would like to express our sincere gratitude to the reviewers for their valuable comments.

\section{Author Contributions}
Z.Z. conceptualized and led the study design, conducted the literature search and data collection, and was responsible for model development and experiments. Z.Z. and M.Z. drafted the initial manuscript. R.Z. provided supervision throughout the study. Z.Z., M.Z., S.Z., Y.S., X.C., Y.H., Y.W., Y.R., and R.Z. participated in research discussions, contributed feedback on methodology and results, reviewed and edited the manuscript, and approved the final version for submission.

\section{Competing Interests}
The authors declare no competing interests.

\section{Funding}
This work was supported by the National Institutes of Health’s National Center for Complementary and Integrative Health (grant numbers R01AT009457 and U01AT012871), the National Institute on Aging (grant number R01AG078154), the National Cancer Institute (grant number R01CA287413), the National Institute of Diabetes and Digestive and Kidney Diseases (grant number R01DK115629), and the National Institute on Minority Health and Health Disparities (grant number 1R21MD019134-01).

\bibliographystyle{unsrt}
\bibliography{0_main}

\end{document}